# KPoEM: A Human-Annotated Dataset for Emotion Classification and RAG-Based Poetry Generation in Korean Modern Poetry

## Authors

**Iro Lim**[1]
The Academy of Korean Studies, Cultural Informatics, Graduate School of Korean Studies
MA Student, Republic of Korea
bkksg.studio@gmail.com

**Haein Ji**[1]
The Academy of Korean Studies, Cultural Informatics, Graduate School of Korean Studies
Ph.D. Student, Republic of Korea
cihayin@gmail.com

**Byungjun Kim**[1]*
The Academy of Korean Studies, Cultural Informatics, Graduate School of Korean Studies
Assistant Professor, Republic of Korea
bjkim@byungjunkim.com

[1]Graduate School of Korean Studies, The Academy of Korean Studies
*Corresponding Author

## Abstract

This study introduces KPoEM (Korean Poetry Emotion Mapping), a novel dataset that serves as a foundation for both emotion-centered analysis and generative applications in modern Korean poetry. Despite advancements in NLP, poetry remains underexplored due to its complex figurative language and cultural specificity. We constructed a multi-label dataset of 7,662 entries (7,007 line-level and 615 work-level), annotated with 44 fine-grained emotion categories from five influential Korean poets. The KPoEM emotion classification model, fine-tuned through a sequential strategy—moving from general-purpose corpora to the specialized KPoEM dataset—achieved an F1-micro score of 0.60, significantly outperforming previous models (0.43). The model demonstrates an enhanced ability to identify temporally and culturally specific emotional expressions while preserving core poetic sentiments. Furthermore, applying the structured emotion dataset to a RAG-based poetry generation model demonstrates the empirical feasibility of generating texts that reflect the emotional and cultural sensibilities of Korean literature. This integrated approach strengthens the connection between computational techniques and literary analysis, opening new pathways for quantitative emotion research and generative poetics. Overall, this study provides a foundation for advancing emotion-centered analysis and creation in modern Korean poetry.

## Acknowledgment

This research was supported by the Academy of Korean Studies (AKS) under Grant No. AKSR2025-RE04 (*Development of Advanced Natural Language Processing and Large Language Model-Based Digital Korean Studies and Education Methodology*, 2025). The authors would like to express their sincere gratitude to the Academy of Korean Studies (AKS) for its technical and financial support, and to Seul Koo, Jonghoon Yun, and Song-yi Jung for their valuable contributions to the data labeling work.

# I. Introduction

Poetry is widely regarded as one of the most expressive forms of literature, capable of capturing subtle nuances of human emotion. Unlike straightforward prose, however, poetic language often conveys feelings indirectly—through metaphor, imagery, and symbolic reference—requiring readers to infer meaning beyond the literal words. This richness of expression makes poetry evocative but also difficult to analyze, even for human readers, and more so for computational models.

Although recent advances in large language models (LLMs) have greatly improved emotion classification in general text, these models often falter when encountering metaphorically dense poetic language. This limitation reveals an interpretive gap, in which the statistical pattern recognition of current AI models fails to adequately capture the subtle emotional expressions embedded in literary texts. In particular, culturally embedded emotional concepts in Korean literature—such as *seoreo-um*(Korean: 서러움; sorrow) and *bijang-ham*(Korean: 비장함; resolute)—further highlight the need for domain-specific resources.

To address this challenge, we developed KPoEM (Korean Poetry Emotion Mapping), the first expert-annotated dataset designed specifically for emotion analysis in modern Korean poetry. For the annotation process, five trained experts provided both line-level and work-level emotional labels, enabling a multi-layered analysis of poetic expression. This human-labeled dataset is, to our knowledge, the first of its kind for Korean literature, and it serves as a crucial foundation for applying and evaluating AI models in this context. KPoEM is constructed from the works of five major modern Korean poets—Han Yong-un(Korean: 한용운), Im Hwa(Korean: 임화), Kim So-wol(Korean: 김소월), Yi Sang(Korean: 이상), and Yun Dong-ju(Korean: 윤동주)—whose writings capture the diverse emotions and nuanced cultural sensibilities of the early twentieth-century Korean poetry (Kim & Cheon, 2020; Seoul Shinmun, 2007).

The purpose of this study is threefold: (1) to construct KPoEM, a specialized dataset for emotion-aware literary computing; (2) to evaluate its utility through a sequential fine-tuning strategy that addresses the complexities of poetic language; and (3) to apply this framework to a RAG-based poetry generation system that reflects culturally grounded Korean emotional sensibilities.

In summary, the key contributions of this paper are as follows:

1. **Introduction of KPoEM:** It presents the first expert-annotated dataset for modern Korean poetry, featuring 7,662 entries with dual-layered (line-level and work-level) emotional labels.

2. **Performance Gain through Sequential Fine-tuning:** It demonstrates that a sequential fine-tuning strategy—transitioning from KcELECTRA[1] pre-trained on general comments to the KOTE(Korean Online That-gul Emotions) dataset (Jeon et al., 2024), and finally to the specialized KPoEM dataset—is highly effective for capturing poetic nuances. This methodological approach resulted in a micro-F1 score of 0.60, representing a significant improvement over the 0.43 baseline and establishing a new

[1] KcELECTRA (Lee, 2021) is a pretrained ELECTRA model (Clark et al., 2020) trained on Korean online news comments. For the development of the KPoEM model, the 2022 version (KcELECTRA-base-v2022) was utilized. See the following repository for details. KcELECTRA, https://github.com/Beomi/KcELECTRA

performance benchmark for emotion analysis in modern Korean poetry.

3. **Bridging Analysis and Generation:** It demonstrates that the KPoEM dataset can be effectively integrated as structured metadata into a vector database within a RAG-based framework. This approach allows for emotion-aware retrieval, enabling the system to produce creative outputs that reflect culturally grounded Korean emotional sensibilities.

Ultimately, this study bridges the gap between natural language processing and the humanities by integrating humanistic insight with computational methodology across dataset construction, model evaluation, and generative modeling. By extending the scope of NLP into figurative and culturally specific domains, our approach enables the systematic investigation of creative questions that have traditionally eluded quantification. This work not only establishes a robust foundation for the large-scale analysis of poetic emotion but also offers new tools for creative exploration, reflecting the core ethos of the digital humanities—where literature can be examined through new lenses without sacrificing its essential contextual nuance.

# II. Background

## A. Emotion Classification in Text-Based Language Models and Literary Texts

Recent advancements in transformer-based LLMs such as BERT (Devlin et al., 2019), RoBERTa (Liu et al., 2019), and GPT-3 (Brown et al., 2020) have significantly improved emotion classification, consistently outperforming earlier methods (Acheampong et al., 2020). When fine-tuned on large, high-quality annotated datasets like GoEmotions (Demszky et al., 2020), these models effectively capture subtle emotional nuances. This confirms that integrating high-quality human annotations with pre-trained LLMs is the most effective approach for accurate text emotion recognition.

However, literary texts pose distinct challenges as metaphorical and stylistic complexities often obscure direct emotional cues. As a result, models trained primarily on literal, contemporary texts frequently misclassify figurative or affectively implicit expressions in literary works (Ji, 2024; S & Mahalakshmi, 2019; Sprugnoli & Redaelli, 2024). To address this interpretive gap, recent scholarship has emphasized the necessity of domain adaptation strategies, including fine-tuning LLMs on corpora composed of literary texts (Li et al., 2021). Zhao et al. (2024), for instance, demonstrated notable advancements in the interpretive capabilities of LLMs by fine-tuning them on a curated corpus of ancient Chinese poetry and incorporating literary features such as stylistic patterns and thematic diversity, underscoring the importance of contextual familiarity with literary forms and conventions.

Empirical studies show that general-purpose sentiment lexicons and binary classification are inadequate for literary analysis due to the affective complexity of works like Horace's Odes (Sprugnoli et al., 2023). This necessitates a shift toward multi-class/multi-label frameworks (Ahmad et al., 2020) and techniques like continual pre-training on literary corpora to capture genre-specific expressions and subtle tones. Though LLMs provide a robust foundation for emotion classification in general-purpose texts, their application to literary materials requires methodological refinement. Integrating domain-specific annotated corpora, adopting multi-dimensional emotion taxonomies, and incorporating insights from literary theory significantly enhance model performance in this complex domain.

## B. Emotion Datasets and Annotation Practices for Korean Language and Poetry

The development of Korean-language emotion datasets has seen notable progress in recent years, laying a crucial foundation for computational analysis of emotional expression in Korean texts. Early lexical resources (Park et al., 2018; Sohn et al., 2012), while providing useful polarity and intensity data, are insufficient for contemporary deep learning. For this reason, GoEmotions-Korean[2] has been developed through the translation and manual correction of the English-language GoEmotions dataset. While this effort expands the availability of Korean emotion-labeled data, scholars caution that directly importing emotion taxonomies from English corpora may overlook culturally embedded emotions unique to Korean language and literature (Jeon et al., 2024).

To address such limitations, large-scale annotated corpora have emerged. The KOTE dataset (Jeon et al., 2024) represents one of the most comprehensive Korean emotion datasets to date, comprising 50,000 online comments with over 250,000 human-annotated labels across 44 emotion categories, including culturally specific emotions. Derived through clustering of emotion terms in embedding spaces, KOTE captures nuanced emotional expressions reflective of Korean sociocultural contexts. Notably, KOTE has been extensively utilized to fine-tune transformer-based models such as KoBERT[3] and KcELECTRA, enabling them to recognize complex, multi-dimensional emotional states beyond simple polarity. However, as KOTE is primarily composed of colloquial online comments, it remains inherently limited in capturing the highly refined, metaphorical, and aesthetically elevated emotional expressions characteristic of literary works. Similarly, Kang's (2024) taxonomy for classical novels addresses narrative-specific needs, yet a gap remains for the metaphorical complexity of poetry.

Despite these advances, poetry remains an underexplored domain because its figurative density, symbolism, and interpretive openness introduce significant ambiguity for annotators. Prior research on poetic emotion annotation provides valuable methodological insights. For instance, international precedents like PO-EMO (Haider et al., 2020) and PERC (Sreeja & Mahalakshmi, 2019) highlight the necessity of expert-led, multi-label annotation to manage the multifaceted nature of poetic effect. However, capturing these subtle nuances remains a significant challenge for computational models—a limitation that recent deep learning approaches have begun to address by effectively modeling long-term dependencies and contextual focus (Ahmad et al., 2020).

Despite the precedents set by KOTE and Kang's taxonomy, the historical depth and layered imagery of early twentieth-century Korean poetry necessitate a more specialized approach to capture its symbolic and culturally unique emotional lexicons. Consequently, constructing an emotion dataset tailored to Korean poetry is vital to bridging the gap between computational linguistics and culturally nuanced AI development.

---

[2]GoEmotions-Korean, https://github.com/monologg/GoEmotions-Korean
[3]KoBERT, https://github.com/SKTBrain/KoBERT

## C. The Advancements of Computational Generative Poetics

Computational generative poetics integrates NLG, Computational Creativity, and AI to produce creative, aesthetically pleasing text. Unlike early knowledge-intensive, symbolic systems like PoeTryMe (Oliveira, 2012) that enforced rigid formal constraints, modern tasks focus on balancing meaningful output with literary nuances (Oliveira, 2017).

Recent advancements utilize pre-trained LLMs and sophisticated decoding. Fine-tuning models like GPT-2 or GPT-Neo on domain-specific corpora has proven superior to general-purpose LLMs in generating emotionally evocative and semantically coherent poems (Bena & Kalita, 2019; Bhat et al., 2025; Lo, 2022). Notably, the TPPoet model balanced diversity and quality through dynamic temperature and Anti-LM decoding[4] (Panahandeh et al., 2023). However, evaluation remains difficult due to the subjectivity of aesthetic judgment and metaphorical complexity (Van Heerden & Bas, 2024). While automatic metrics provide baseline validation, human assessment of poeticness and coherence remains indispensable (Panahandeh et al., 2023).

This trajectory suggests that domain-specific emotion resources are essential for building culturally grounded poetry generation models. Heflin (2020) argues that AI-generated literature is not a monolithic practice but rather a complex assemblage of human and machine labor, emphasizing the role of vector representations in making language legible to generative models. Because emotional expression in modern Korean poetry is highly metaphorical and symbolic, the vectorization of nuanced affective states becomes a prerequisite for effective computational modeling. Recent studies therefore highlight the need for emotion-aware representations that can support faithful modeling of complex literary affect beyond surface-level sentiment.

[4] The Anti-LM is a contrastive decoding strategy designed to suppress prior bias in Large Language Models (LLMs), particularly the tendency to repeat source text or follow repetitive patterns in zero-shot settings. By penalizing the probabilities (logits) of a simple language model conditioned only on the source, this method encourages more diverse and instruction-aligned generation. (For the technical formulation involving exponential decay, see Sia et al., 2024).

# III. Methodology

## A. Overview of Dataset Construction[5]

This study constructed KPoEM (Korean Poetry Emotion Mapping), an emotion-annotated dataset for quantitative analysis and emotion classification modeling of modern Korean poetry. The dataset was constructed through a structured preprocessing and refinement workflow, detailed in Appendix A, including poet and text selection, text normalization, and emotion annotation. Five expert annotators assigned multiple emotion labels to each instance based on the 44 fine-grained emotion categories defined in KOTE. Emotion categories are ordered alphabetically by their Korean labels for neutrality and ease of reference (see Table 1). Designed for transformer-based models, KPoEM supports modeling emotional complexity in modern Korean poetry and downstream DH and NLP tasks. Examples of the finalized dataset are presented in Table 2 and Table 3.[6]

**Table 1. Emotion Categories (n = 44) Used for KPoEM (Based on KOTE[7])**

| Valence | Korean | Romanized | Interpretation |
|---|---|---|---|
| **Negative** | 경악 | *gyeongak* | shock |
| | 공포/무서움 | *gongpo/museoum* | fear |
| | 귀찮음 | *gwichaneum* | laziness |
| | 당황/난처 | *danghwang/nancheo* | embarrassment |
| | 부끄러움 | *bukkeureoum* | shame |
| | 부담/안 내킴 | *budam/an naekim* | reluctant |
| | 불쌍함/연민 | *bulssangham/yeonmin* | compassion |
| | 불안/걱정 | *buran/geokjeong* | anxiety |
| | 불평/불만 | *bulpyeong/bulman* | dissatisfaction |
| | 슬픔 | *seulpeum* | sadness |
| | 서러움 | *seoreoum* | sorrow |
| | 안타까움/실망 | *antakkaum/silmang* | disappointment |
| | 어이없음 | *eoieopseum* | preposterous |
| | 역겨움/징그러움 | *yeokgyeoum/jinggeureoum* | disgust |

[5] The KPoEM dataset constructed in this research is available at the following link. https://doi.org/10.57967/hf/6303

[6] The English translations provided in parentheses within the *text*, *sub_title*, and *title* columns in Table 2 and Table 3 were generated using Gemini 3 Pro developed by Google, and subsequently reviewed and validated by the authors.

[7] The valence classification and English interpretations of the emotion categories are adopted from the original KOTE (Korean Online That-gul Emotions) dataset, following Jeon et al. (2024). The dataset is publicly available at https://github.com/searle-j/KOTE

| | | | |
|---|---|---|---|
| | 의심/불신 | *uisim/bulsin* | distrust |
| | 짜증 | *jjajeung* | irritation |
| | 재미없음 | *jaemieopseum* | boredom |
| | 절망 | *jeolmang* | despair |
| | 죄책감 | *joechaekgam* | guilt |
| | 증오/혐오 | *jeungo/hyeomo* | contempt |
| | 지긋지긋 | *jigeutjigeut* | fed up |
| | 패배/자기혐오 | *paebae/jagihyeomo* | gessepany |
| | 한심함 | *hansimham* | pathetic |
| | 화남/분노 | *hwanam/bunno* | anger |
| | 힘듦/지침 | *himdeum/jichim* | exhaustion |
| **Positive** | 감동/감탄 | *gamdong/gamtan* | admiration |
| | 고마움 | *gomaum* | gratitude |
| | 기대감 | *gidaegam* | expectancy |
| | 기쁨 | *gippeum* | joy |
| | 뿌듯함 | *ppudeutam* | pride |
| | 신기함/관심 | *singiham/gwansim* | interest |
| | 아껴주는 | *akkyeojuneun* | care |
| | 안심/신뢰 | *ansim/silloe* | relief |
| | 존경 | *jongyeong* | respect |
| | 즐거움/신남 | *jeulgeoum/sinnam* | excitement |
| | 편안/쾌적 | *pyeonan/kwaejeok* | comfort |
| | 행복 | *haengbok* | happiness |
| | 환영/호의 | *hwanyeong/houi* | welcome |
| | 흐뭇함(귀여움/예쁨) | *heumutam(gwiyeoum/yeppeum)* | attracted |
| **Neutral** | 깨달음 | *kkaedareum* | realization |
| | 놀람 | *nollam* | surprise |

|  | 비장함 | *bijangham* | resolute |
|---|---|---|---|
|  | 우쭐댐/무시함 | *ujjuldaem/musiham* | arrogance |
| **ETC.** | 없음 | *eopseum* | NO EMOTION |

**Table 2. Examples from the KPoEM Line-Level Dataset**

| **line_id** | **poem_id** | **text** | **sub_title** | **title** | **poet** | **annotator_01** | **annotator_02** | **annotator_03** | **annotator_04** | **annotator_05** |
|---|---|---|---|---|---|---|---|---|---|---|
| 36 | 4 | 바람이 부는데 (While the wind blows) |  | 바람이 불어 (In the Blowing Wind) | Yun Dong-ju | interest, NO EMOTION | interest, NO EMOTION, embarrassment,anxiety, resolute, sorrow | NO EMOTION | anxiety, reluctant, boredom | admiration, joy, surprise, comfort, welcome |
| 6885 | 472 | 꽃이보이지않는다. (No flowers are in sight.) | 절벽 (Precipice) | 위독 (Critical Condition) | Yi Sang | despair, disappointment, fear, anxiety, embarrassment, reluctant, distrust | fear, embarrassment, anxiety, shock, sorrow, disappointment, despair, distrust, realization | embarrassment, disappointment | shock, fear, embarrassment, reluctant, sadness, despair | embarrassment, anxiety, disappointment |

**Table 3. Examples from the KPoEM Work-Level Dataset**

| seg_id | poem_id | text | sub_title | title | poetry_book | poet | annotator_01 | annotator_02 | annotator_03 | annotator_04 | annotator_05 |
|---|---|---|---|---|---|---|---|---|---|---|---|
| 6 | 6 | 계절이 지나가는<br>하늘에는<br>가을로 가득 차 있습니다.<br><br>나는 아무 걱정도 없이<br>가을 속의 별들을 다 헤일<br>듯합니다...<br><br>가슴 속에 하나 둘<br>새겨지는 별을<br>이제 다 못 헤는 것은<br>쉬이 아침이 오는<br>까닭이요,<br>내일 밤이 남은 까닭이요,<br>아직 나의 청춘이 다하지<br>않은 까닭입니다.<br><br>별 하나에 추억과<br>별 하나에 사랑과<br>별 하나에 쓸쓸함과<br>별 하나에 동경과<br>별 하나에 시와<br>별 하나에 어머니, 어머니<br>(The sky where seasons<br>pass by<br>is filled with autumn.<br><br>I feel as though I could<br>count<br>all the stars in the autumn<br>air, without a single care...<br><br>The reason I cannot count<br>all the stars<br>being engraved one by one<br>in my heart now<br>is because morning comes<br>too soon,<br>because tomorrow night<br>still remains,<br>and because my youth is<br>not yet spent.<br><br>Memory for one star,<br>Love for another,<br>Loneliness for another, | | 별 헤는 밤 (The Night Counting Stars) | 하늘과 바람과 별과 시 (The Sky, the Wind, the Stars, and the Poem) | Yun Dong-ju | attracted, admiration, care, sadness, joy, expectancy, realization, welcome, respect | admiration, expectancy, sorrow, sadness, resolute, care, attracted, disappointment, realization, respect | attracted, expectancy, disappointment, sorrow, sadness | admiration, joy, attracted, happiness, care, comfort | respect, admiration, interest, realization, sorrow, sadness, disappointment |

| | | Longing for another,<br>Poetry for another,<br>And mother, mother for<br>one star.) | | | | | | | | | |
|---|---|---|---|---|---|---|---|---|---|---|---|
| **7** | **6** | 어머님, 나는 별 하나에<br>아름다운 말 한 마디씩<br>불러 봅니다. 소학교 때<br>책상을 같이했던<br>아이들의 이름과, 패, 경,<br>옥 이런 이국 소녀들의<br>이름과, 벌써 아기 어머니<br>된 계집애들의 이름과,<br>가난한 이웃 사람들의<br>이름과, 비둘기, 강아지,<br>토끼, 노새, 노루,<br>'프랑시스 잠', '라이너<br>마리아 릴케', 이런 시인의<br>이름을 불러 봅니다.<br><br>이네들은 너무나 멀리<br>있습니다.<br>별이 아스라이 멀 듯이,<br><br>어머님,<br>그리고 당신은 멀리<br>북간도에 계십니다<br><br>나는 무엇인지 그리워<br>이 많은 별빛이 내린 언덕<br>위에<br>내 이름자를 써 보고,<br>흙으로 덮어<br>버리었습니다.<br><br>딴은, 밤을 새워 우는<br>벌레는<br>부끄러운 이름을<br>슬퍼하는 까닭입니다.<br><br>그러나 겨울이 지나고<br>나의 별에도 봄이 오면<br>무덤 위에 파란 잔디가<br>피어나듯이<br>내 이름자 묻힌 언덕<br>위에도<br>자랑처럼 풀이 무성할<br>게외다. (Mother, I call out a beautiful word for every star. The names of the children I shared a desk with in elementary school; the names of foreign girls like Pae, Kyeong, and Ok; the names of girls who have already become mothers; the names of poor | | 별 헤는 밤 (The Night Counting Stars) | 하늘과 바람과 별과 시 (The Sky, the Wind, the Stars, and the Poem) | Yun Dong-ju | care, sadness, shame, compassion, respect, admiration, gratitude, attracted, welcome | care, attracted, welcome, sorrow, sadness, realization, guilt, shame, compassion, expectancy | disappointment, sorrow, expectancy, sadness, realization | anxiety, sorrow, sadness, respect, welcome | sorrow, sadness, respect |
| | | neighbors; and the names of poets like 'Francis Jammes' and 'Rainer Maria Rilke', along with pigeons, puppies, rabbits, mules, and roe deer.<br><br>They are all so far away. As distantly far as the stars,<br><br>Mother, And you are far away in Northern Kando.<br><br>Longing for I know not what, I wrote my name upon this hill where so much starlight falls, And then I covered it over with dirt.<br><br>Surely, the reason the insects chirp all through the night is because they grieve for their shameful names.<br><br>But when winter passes and spring comes to my star as well, Just as green grass sprouts upon a grave, Grass will grow thick like pride upon the hill where my name is buried.) | | | | | | | | | |

In this study, the KPoEM dataset was constructed in two forms: line-level and work-level. For the line-level dataset, poetry texts were cleaned and segmented into individual lines, with a portion of the data randomly shuffled to allow annotators to focus on line-specific emotional expression without broader contextual influence. This design enables an experimental examination of whether individual lines can convey emotions independently of their surrounding context. The dataset was constructed from 483 poems following the metadata schema shown in Table 4.

To account for the contextual nature of poetic emotion, a work-level dataset was also created in which annotators read each poem in its entirety and assigned emotion labels with full contextual awareness. Each poem was treated as a single data instance; however, texts exceeding 512 characters were segmented into paragraphs and ordered sequentially according to the metadata schema in Table 5. The original line and stanza structures from Wikisource were preserved, and emotion annotation was performed on texts that retained these poetic formal structures, ensuring that the formal and rhythmic characteristics of the source poems were reflected in the annotation process.

**Table 4. Line-Level Metadata Schema of KPoEM**

| Field Name | Description |
|---|---|
| **line_id** | Unique identifier for each line-level entry in the dataset |
| **poem_id** | Individual identifier assigned to each poem included in the dataset |
| **text** | Text content of the individual poetic line. |
| **sub_title** | Subtitle of an individual piece in a series (if applicable) (e.g., Pursuit) |
| **title** | Title of the poem (e.g., Azaleas) |
| **poet** | The author of the poem (e.g., Kim So-wol) |
| **annotator_XX** | Identifier for the person or group who annotated the given line (e.g., annotator_01) |

**Table 5. Work-Level Metadata Schema of KPoEM**

| Field Name | Description |
|---|---|
| **seg_id** | Unique identifier for each work-level entry in the dataset |
| **poem_id** | Individual identifier assigned to each poem included in the dataset |
| **text** | Full text of the poem |
| **sub_title** | Subtitle of an individual piece in a series (if applicable) (e.g., Sinking) |
| **title** | Title of the poem (e.g., In the Blowing Wind) |
| **poetry_book** | Title of the poetry collection in which the poem appears (e.g., Sky, Wind, Stars, and Poetry) |
| **poet** | Name of the poet (e.g., Han Yong-un) |
| **annotator_XX** | Identifier for the person or group who annotated the given work (e.g., annotator_05) |

In both dataset types, a multi-label structure was adopted, allowing five annotators to assign up to ten emotion labels to each line (or work). The order of metadata fields was designed so that annotators would first encounter the poem text immediately after the ID field, followed by the title and author. For series-based poems (e.g., Yi Sang's *Widok*), a subtitle field was added to distinguish between individual pieces, such as *Chugu* (Korean: 추구; Pursuit) and *Chimmol* (Korean: 침몰; Sinking).

Table 6 presents the number of poetic lines and poems per poet in KPoEM. In total, the dataset comprises a total of 7,622, consisting of 7,007 line-level segments and 615 work-level texts drawn from 483 representative works of the 1920s–1940s. An analysis of the emotion label distribution revealed the ten most frequently assigned categories (see Table 7). Beyond its role as an analytical corpus, this dataset serves as the foundation for emotion-conditioned poetry generation experiments, enabling the modeling of nuanced affective transitions in modern Korean literature.

**Table 6. Summary of the Number of Poetic Lines and Poems Per Poet in KPoEM**

| Category | Han Yong-un | Im Hwa | Kim So-wol | Yi Sang | Yun Dong-ju | Total |
|---|---|---|---|---|---|---|
| **Line-level** | 1,198 | 2,163 | 2,071 | 464 | 1,111 | 7,007 |
| **Work-level** | 138 | 110 | 176 | 77 | 114 | 615 |
| **Subtotal** | | | | | | **7,662** |
| **Number of Poems** | **117** | **43** | **165** | **46** | **112** | **483** |

**Table 7. Top 10 Most Frequently Used Emotion Labels**

| Rank | Emotion Label | Frequency |
|---|---|---|
| 1 | anxiety | 10,126 |
| 2 | sadness | 8,715 |
| 3 | expectancy | 8,399 |
| 4 | disappointment | 8,016 |
| 5 | sorrow | 7,423 |
| 6 | interest | 7,094 |
| 7 | resolute | 6,808 |
| 8 | care | 6,786 |
| 9 | admiration | 5,316 |
| 10 | embarrassment | 5,249 |

Table 8 presents statistics on inter-annotator agreement for emotion labels. Across the dataset, 99% of the texts show agreement by at least two annotators on one or more emotion labels, indicating that no major disagreements occurred during the annotation process. This high level of agreement reflects not only the effectiveness of the annotation guidelines but also the shared cultural and linguistic interpretive context among annotators. In addition, texts labeled as NO EMOTION exhibit a balanced distribution (see Table 9), acknowledging that not all poetic lines explicitly convey emotion. This distribution enhances the dataset's representativeness of real reading experiences and provides a foundation for training models to distinguish emotionally weak or absent expressions.

**Table 8. Statistics of Inter-Annotator Agreement on Emotion Labels**

| Agreement | | | | | |
|---|---|---|---|---|---|
| **at least one label of x or higher** | **x=1** | **x=2** | **x=3** | **x=4** | **x=5** |

| # of texts<br>(% to total) | 7,622 | 7,613 | 7,052 | 4,659 | 1,725 |
|---|---|---|---|---|---|
| | 100 | 99.88 | 92.52 | 61.13 | 22.63 |

**Table 9. Statistics of Lines Labeled as NO EMOTION**

| Texts Labeled for NO EMOTION | | | | | | |
|---|---|---|---|---|---|---|
| **NO EMOTION** | **0** | **1** | **2** | **3** | **4** | **5** |
| **# of texts<br>(% to total)** | 6,047<br>(79.34%) | 897<br>(11.77%) | 413<br>(5.42%) | 208<br>(2.73%) | 50<br>(0.66%) | 7<br>(0.09%) |

The KPoEM dataset was made publicly available via Zenodo and Hugging Face. During distribution, the shuffled line-level dataset was reordered according to the original 'line_id' sequence. For the work-level dataset—which preserves the structural features of poems from Wikisource and was annotated in that form—the release version was standardized by removing newline characters (\n) and retaining only the continuous text. This adjustment was made to enhance the consistency and usability of the dataset, as unnecessary newline symbols can introduce unintended effects during input tokenization in model training and comparative experiments.

## B. Model Construction[8]

### 1. TVT distribution

For model training and evaluation, we established a rigorous data partitioning protocol. The procedure began by handling the two distinct data formats—the poem work-level dataset and the poem line-level dataset—separately. First, each of the two datasets was independently split into training, validation, and test sets with an 8:1:1 distribution (See Table 10). Following this initial split, the corresponding sets were merged: the training set from the work-level data was combined with the training set from the line-level data, and this process was repeated for the validation and test sets to yield the final, unified datasets.

**Table 10. Distribution of Data across TVT (Training, Validation, Test) Sets**

| | Train (80%) | Validation (10%) | Test (10%) | Total |
|---|---|---|---|---|
| **The Number of Rows** | 6,096 | 763 | 763 | 7,622 |

### 2. Processing of Multi-Annotator Data for Fine-Tuning

The process for handling multi-annotator data involves three main steps: aggregation, vectorization, and normalization.

a. **Label Aggregation**: For each data instance, all discrete labels assigned by the 5 annotators are

[8] The KPoEM emotion classification model constructed in this research is available at the following link. https://doi.org/10.57967/hf/6303

collected and compiled into a comprehensive bag of labels, ensuring all annotations (annotator_01 to annotator_05) are preserved.

b. **Score Vectorization**: The aggregated labels are transformed into a numerical 'score vector' ($s$). This vector's dimension is equal to the total number of possible emotions ($L$). Each entry in s represents the frequency (or 'vote count', 0 to 5) of a specific emotion from $L$ across all annotators, thereby quantifying the level of consensus for each label.

c. **Normalization**: To account for varying levels of agreement across instances, the raw score vector is normalized on an instance-by-instance basis using Min-Max scaling. This rescales the scores to a continuous range between 0 and 1, emphasizing the relative importance of each emotion within that specific data instance.

The foundation model used for classification was KcELECTRA-base-v2022, initially fine-tuned on the KOTE dataset. An additional fine-tuning process was then performed using the KPoEM training data, with validation data employed for monitoring progress and mitigating overfitting.

# IV. Results

Model performance was evaluated using the test set (n = 763), applying a classification threshold of 0.30. The threshold of 0.30 indicates that, among the 44 emotion categories, an emotion label is regarded as correctly predicted when its score exceeds 0.30. This standard directly adopts the threshold proposed in previous research by Jeon et al. (2024).

Table 11 presents a comparative analysis of the performance of models trained on the KPoEM and KOTE datasets, respectively. For this research phase, we utilized Optuna (v4.6.0), a hyperparameter optimization framework, to identify the optimal values for key hyperparameters within a Python (v3.12.3) environment. The search was conducted for 3 epochs within the following ranges: a learning rate between 1e-6 and 5e-5, a batch size between 8 and 16, and a dropout rate between 0.1 and 0.5. The resulting emotion probability distributions were normalized using min-max scaling (min=0, max=1), and the final labels were derived by applying a classification threshold of 0.3, subsequent to the data preprocessing which also involved a min-max scaling factor of 0.2.

The KcELECTRA model, when subjected to direct fine-tuning solely on the KOTE dataset[9], exhibited relatively low performance, achieving an Accuracy of 0.77, a micro-averaged Recall of 0.38, and a macro-averaged F1-score of 0.34. This outcome suggests that the KOTE dataset, which primarily comprises emotion data from colloquial online language, has limitations in capturing the nuanced contextual emotions inherent in literary texts. In contrast, the model directly fine-tuned on our KPoEM dataset[10] recorded a superior performance with an Accuracy of 0.79 and a macro F1-score of 0.45. This result demonstrates that the KPoEM dataset provides a stable and effective foundation for the task of emotion classification in Korean modern

---

[9] The KcELECTRA model fine-tuned only on the KOTE dataset is available at the following link. https://huggingface.co/AKS-DHLAB/KcELECTRA_KOTEOnly

[10] The KcELECTRA model fine-tuned only on the KPoEM dataset is available at the following link. https://huggingface.co/AKS-DHLAB/KcELECTRA_KPoEMOnly

poetry.

Notably, the model employing a sequential fine-tuning approach—pre-training on the KOTE dataset before transfer learning on the KPoEM dataset—yielded the best performance across all metrics: an Accuracy of 0.79, a micro-averaged Recall of 0.69, a macro F1-score of 0.49, and an MCC of 0.47. This finding indicates that a sequential training strategy is highly effective for literary emotion classification, significantly improving the balance between precision and recall as reflected by the F1-score. Our experiments, leveraging optimized hyperparameters from the Optuna search, revealed the superior performance of the sequential fine-tuning model over single-dataset approaches. This finding implies that for domains with limited data, such as literary emotion analysis, the supplementary use of a broader, general-domain emotion dataset is a highly effective strategy for improving model performance.

**Table 11. Performance Comparison of KPoEM Emotion Classification Models (Threshold = 0.3)**

| Model | Accuracy | Precision_micro | Precision_macro | Recall_micro | Recall_macro | F1_micro | F1_macro | MCC |
|---|---|---|---|---|---|---|---|---|
| **KcELECTRA (KOTE only)** | 0.77 | 0.49 | 0.46 | 0.38 | 0.33 | 0.43 | 0.34 | 0.29 |
| **KcELECTRA (KPoEM only)** | 0.79 | 0.53 | 0.43 | 0.66 | 0.50 | 0.59 | 0.45 | 0.45 |
| **KcELECTRA (KOTE → KPoEM)** | **0.79** | **0.53** | **0.47** | **0.69** | **0.54** | **0.60** | **0.49** | **0.47** |

To further examine model behavior qualitatively, emotion classification was applied to a selection of representative modern and contemporary Korean poems. The sample comprises works by eminent poets, notably Han Kang (Korean: 한강), laureate of the 2024 Nobel Prize in Literature; canonical figures including Jeong Ji-yong (Korean: 정지용). Comparative analysis revealed distinct tendencies between the three models. Table 12 presents the results of a qualitative evaluation conducted by applying the models trained on the KPoEM and KOTE datasets to actual poetic texts. For this purpose, Han Kang's *Hyoege. 2002. Gyeoul* (Korean: 효에게. 2002. 겨울; To Hyo: Winter 2002) (Han, 2013) and Jeong Ji-yong's *Hyangsu* (Korean: 향수; Nostalgia)[11] were analyzed as case studies.[12]

[11] For the original Korean text of Jeong Ji-yong's *Hyangsu*, we referred to Wikisource (https://ko.wikisource.org/wiki/향수/향수).

[12] The English translations provided in parentheses within the *Text*, *Title* columns in Table 12 were generated using Gemini 3 Pro, and subsequently reviewed and validated by the authors.

**Table 12. Comparative Emotion Classification Results for Poems by Han Kang and Jeong Ji-yong**

| Text | Title | Poet | KcELECTRA (KOTE only) | KcELECTRA (KPoEM only) | KcELECTRA (KOTE → KPoEM) |
|---|---|---|---|---|---|
| 하지만 곧<br>너도 알게 되겠지<br>내가 할 수 있는 일은<br>기억하는 일뿐이란 걸<br>저 번쩍이는 거대한<br>흐름과<br>시간과<br>성장(成長),<br>집요하게 사라지고<br>새로 태어나는 것들 앞에<br>우리가 함께 있었다는 걸<br>(But soon<br>You too will come to realize<br>That the only thing I can do<br>Is to remember<br>That flashing, immense<br>flow, and<br>Time, and<br>Growth,<br>Before things that<br>relentlessly vanish<br>And are born anew<br>That we were together) | 효에<br>게.<br>2002.<br>겨울<br>(To<br>Hyo:<br>Winter<br>2002) | Han<br>Kang | realization: 0.80<br>expectancy: 0.58<br>resolute: 0.55<br>disappointment: 0.51<br>sadness: 0.48<br>anxiety: 0.45<br>NO EMOTION: 0.42<br>exhaustion: 0.39<br>despair: 0.32 | resolute: 0.89<br>realization: 0.86<br>sorrow: 0.77<br>anxiety: 0.72<br>disappointment: 0.69<br>sadness: 0.67<br>expectancy: 0.63<br>exhaustion: 0.42<br>admiration: 0.42<br>dissatisfaction: 0.33<br>distrust: 0.33<br>reluctant: 0.32 | realization: 0.93<br>resolute: 0.91<br>expectancy: 0.75<br>anxiety: 0.50<br>admiration: 0.48<br>sadness: 0.45<br>**relief: 0.44**<br>disappointment: 0.43<br>sorrow: 0.39<br>joy: 0.39<br>**welcome: 0.36**<br>**care: 0.35**<br>**pride : 0.32** |
| 흙에서 자란 내 마음<br>파아란 하늘 빛이 그립어<br>함부로 쏜 화살을 찾으려<br>풀섶 이슬에 함추름<br>휘적시든 곳,<br><br>— 그 곳이 참하 꿈엔들<br>잊힐 리야.<br><br>전설바다에 춤추는 밤불결<br>같은<br>검은 귀밑머리 날리는<br>어린 누이와<br>아무렇지도 않고 여쁠<br>것도 없는<br>사철 발벗은 안해가<br>따가운 해ㅅ살을 등에<br>지고 이삭 줏던 곳,<br><br>— 그 곳이 참하 꿈엔들 | 향수<br>(Nosta<br>lgia) | Jeon<br>g<br>Ji-yo<br>ng | sadness: 0.91<br>disappointment: 0.72<br>compassion: 0.66<br>despair: 0.56<br>exhaustion: 0.54<br>sorrow: 0.49<br>anxiety: 0.42<br>realization: 0.35<br>NO EMOTION: 0.31 | disappointment: 0.94<br>sorrow: 0.94<br>sadness: 0.94<br>anxiety: 0.85<br>care: 0.79<br>compassion: 0.78<br>exhaustion: 0.68<br>realization: 0.63<br>expectancy: 0.59<br>interest: 0.51<br>admiration: 0.40<br>embarrassment: 0.38<br>reluctant: 0.34<br>dissatisfaction: 0.33 | sadness: 0.97<br>**sorrow: 0.94**<br>disappointment: 0.90<br>**compassion: 0.75**<br>anxiety: 0.68<br>care: 0.64<br>exhaustion: 0.63<br>expectancy: 0.42<br>**despair: 0.34**<br>**realization: 0.32**<br>reluctant: 0.31 |

| 잊힐 리야.<br><br>(My mind, raised from the soil<br>Longing for the blue sky light<br>To find the arrow I shot at random<br>Where I was drenched in the dew of the grass thickets,<br><br>— Could that place ever be forgotten, even in dreams?<br><br>Like the night waves dancing in the sea of legend<br>With my young sister, her black side-locks flying<br>And my wife, neither extraordinary nor pretty,<br>Barefoot throughout the four seasons<br>Gleaning ears of grain, the stinging sunlight on her back<br><br>— Could that place ever be forgotten, even in dreams? | | | | | |
|---|---|---|---|---|---|

In a case study of Han Kang's poem, the KOTE-only model struggled with poetic nuances, predicting less contextually aligned emotions such as 'despair.' In contrast, the KPoEM-only model showed improved alignment with 'realization' and 'sorrow,' though with residual noise. The transfer learning model (KOTE → KPoEM) achieved the most stable distribution, identifying deeper existential solidarity (e.g., relief, welcome) beyond the tragic tone.

In a case study of Jeong Ji-yong's *Hyangsu*, the KOTE-only model exhibited a diffuse emotional distribution, overemphasizing less contextually appropriate emotions such as despair and exhaustion. In contrast, the KPoEM-only model showed improved alignment with core emotions of longing and compassion, though with residual noise due to uneven emotional concentration. The transfer learning model (KOTE → KPoEM) achieved the most stable emotional hierarchy, effectively suppressing excessive negative affect while capturing relational and existential dimensions of memory and belonging beyond the poem's tragic tone. These findings indicate that general-domain pretraining followed by domain-specific refinement achieves the best balance between emotional intensity and semantic alignment in literary emotion classification. We refer to this optimized transfer model as the KPoEM emotion classification model.

# V. Emotion-Aware Generative Poetics with KPoEM

## A. Overview: Emotion-Aware Poetry Generation Framework

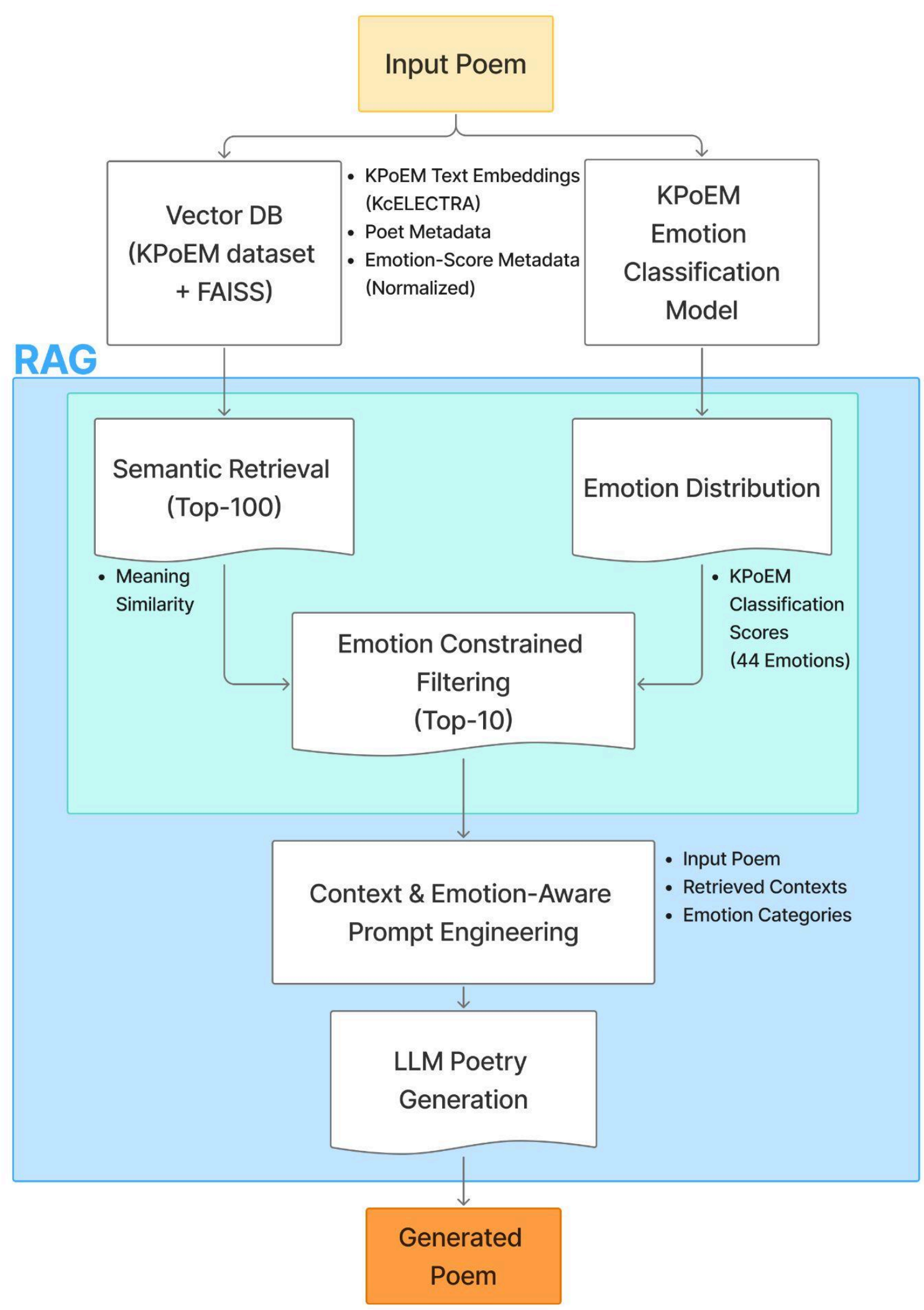

**Figure 1. Emotion-Aware RAG-Based Poetry Generation Architecture**

This section presents an emotion-aware poetry generation framework based on Retrieval-Augmented Generation (RAG), which integrates semantic similarity with emotion classification using the KPoEM dataset. The system generates poems by retrieving semantically relevant and emotionally aligned poetic lines and using them as contextual input for a large language model (LLM). As illustrated in Figure 1, the framework consists of emotion classification, emotion-constrained filtering, and LLM-based generation. The following sections provide step-by-step examples demonstrating how this pipeline operates in practice.[13]

## B. Poetry Generation Experiments and Results

### 1. Construction of the Vector Database

This study constructs a vector database from the KPoEM dataset to support RAG-based poetry generation. In this framework, the vector database acts as a specialized external memory that provides the LLM with domain-specific knowledge of modern Korean poetic sensibilities (Jing et al., 2025). Each poetic line is embedded using the KcELECTRA model, which is also employed as the backbone of the KPoEM emotion classifier, ensuring consistency between semantic representation and emotion interpretation. Each line is associated with normalized emotion-score metadata derived from annotations by five expert annotators. Emotion scores are min–max normalized to the 0–1 range, and only categories with values of 0.2 or higher are retained, following the same thresholding strategy used in the KOTE schema and the KPoEM model. This procedure ensures that the stored metadata reflects the relative emotional salience of each poetic line.

Through this design, the retrieval module selects poetic lines based on three criteria:

(1) semantic similarity to the input text,

(2) emotional alignment within the 44-category KPoEM taxonomy, and

(3) poet-specific contextual patterns.

The resulting vector database is indexed using FAISS (Facebook AI Similarity Search, v1.13.2) and integrated into a LangChain (v1.2.0) pipeline, enabling retrieved poetic lines to be incorporated directly as contextual input during poem generation. The implementation relies on langchain-core (v1.2.5) and langchain-community (v0.4.1) modules and langchain-huggingface (v1.2.0).

### 2. Pipeline Design and Execution

#### 2.1. Emotion Classification of the Input Sentence

The provided input poem is first analyzed by the KPoEM emotion classification model. This model generates probability-based scores across 44 fine-grained emotion categories (See Table 1). These scores serve as reference values for computing the emotional alignment between the input poem and the poetic instances stored in the vector database during the subsequent retrieval stage.

Table 13 presents an example of emotion scores produced by the KPoEM classifier when an excerpt from Kim Chunsu's poem, *Kkot*(Korean: 꽃; Flower) (Kim, 2004), is used as the input text. Kim Chunsu's poetry is an appropriate case for this study, given that his concept of *Ingong-sihak* (Korean: 인공시학; artificial poetics) has been identified as forming an early lineage of Korean AI generative poetics (Jeong, 2025). As shown in the table, the input text exhibits high scores for emotion categories such as Care (0.90), Realization

[13] The English translations of the input poem and the generated poetic output presented in this section were produced using Gemini 3 Pro, and subsequently reviewed and validated by the authors.

(0.87), and Admiration (0.86).

The proposed poetry generation model leverages these classification results by using the input poem as a source of poetic imagery and thematic cues, while relying on the identified emotion categories to determine the affective tone of the generated poem, thereby aligning generation with both the input text and the emotional structures encoded in the KPoEM dataset.

**Table 13. Example of KPoEM Emotion Classification Results for the Poem by Kim Chunsu**

| Original Input Poem | KPoEM Emotion Classification Results |
|---|---|
| 내가 그의 이름을 불러주기 전에는<br>그는 다만<br>하나의 몸짓에 지나지 않았다.<br><br>내가 그의 이름을 불러 주었을 때<br>그는 나에게로 와서<br>꽃이 되었다.<br>(Before I called his name<br>He was nothing<br>But a mere gesture.<br><br>When I called his name<br>He came to me<br>And became a flower.) | care: 0.90<br>realization: 0.87<br>admiration: 0.86<br>joy: 0.79<br>expectancy: 0.78<br>resolute: 0.78<br>welcome: 0.64<br>gratitude: 0.64<br>sadness: 0.63<br>happiness: 0.62<br>respect: 0.60<br>relief: 0.60<br>pride: 0.54<br>disappointment: 0.52<br>sorrow: 0.59<br>attracted: 0.47<br>interest: 0.32<br>anxiety: 0.30 |

## 2.2. Semantic Retrieval and Emotion-Based Filtering

The embedding vector of the input poem is compared against the poetic line vectors stored in the KPoEM vector database. Based on cosine similarity, the system first retrieves the top 100 semantically similar poetic lines—a candidate pool size determined through empirical experimentation to ensure sufficient thematic diversity. It then computes affective similarity by comparing the emotion scores of the input text with the emotion metadata attached to each retrieved line. Subsequently, the system selects the top 10 poetic lines that satisfy both semantic similarity and emotional alignment. This two-stage filtering strategy is designed to mitigate the performance degradation caused by irrelevant context in RAG systems (Leto et al., 2024). By limiting the context to the top 10 emotionally aligned lines, we aim to maximize generation quality while maintaining emotional coherence, consistent with findings that optimal RAG performance is often observed with a context size of around 10 documents (Leto et al., 2024).

**Table 14. Examples of Poetic Lines Retrieved from the KPoEM Vector Database Based on Semantic and Affective Similarity to Flower by Kim Chunsu**

| Poetic Line | Emotion Scores (Top 3) | Poet |
|---|---|---|
| 혀끝에서 물결이 솟고 붓 아래에 꽃이 피어요.<br>(Waves rise from the tip of the tongue, and flowers bloom beneath the brush.) | admiration: 0.8<br>attracted: 0.8<br>joy: 0.8 | Han Yong-un |
| 인간(人間)에 이 세상에 다시 있으랴.<br>(Could such a person ever be found in this world again?) | resolute: 0.6<br>realization: 0.6<br>admiration: 0.4 | Kim So-wol |
| … | | |
| 홈싹홈싹 숨치우는 보드라운 모래 바닥과 같은 긴 길이,<br>항상 외롭고 힘없는 저의 발길을 그리운 당신한테로<br>인도하여 주겠지요.<br>(A long path, like a soft bed of sand breathing with gentle gasps, will surely lead my always lonely and feeble footsteps toward you, for whom I long.) | sadness: 0.8<br>expectancy: 0.8<br>gratitude: 0.6 | Kim So-wol |

Table 14 presents the top ten poetic lines selected from the vector database when Kim Chun-su's poem Flower is used as the input text. As shown in the table, the retrieved lines are selected based not only on semantic similarity but also on emotional alignment.

## 2.3. Prompt Engineering and Poetry Generation

The final set of ten retrieved poetic lines is provided as contextual input to the generative language model (LLM). In this study, Midm-2.0-Base-Instruct, an open-source Korean LLM, was employed, and the retrieved lines and emotion information were dynamically injected into the prompt using the LangChain framework. The generation pipeline was implemented using a LangChain-based pipeline (LangChain v1.2.0), with the following hyperparameters which were determined through empirical experimentation: temperature=0.7, top_p=0.9, max_new_tokens=128, and repetition_penalty=1.2.

Under this configuration, the model generates emotionally coherent poetic text by jointly referencing the retrieved lines and the emotion vector derived from the input. The prompt instruction used in this generation process is provided in Appendix B.

**Table 15. Top Five Emotion Categories and Scores for the Input Poem and the Generated Poem**

| Input Poem | Generated Poem |
|---|---|

| 내가 그의 이름을 불러주기 전에는<br>그는 다만<br>하나의 몸짓에 지나지 않았다.<br><br>내가 그의 이름을 불러 주었을 때<br>그는 나에게로 와서<br>꽃이 되었다.<br>(**Before I called his name**<br>He was nothing more<br>But a mere gesture.<br><br>When I called his name<br>He came to me<br>**And became a flower.**) | <이름 부르는 순간><br><br>나 그대를 부르기 이전엔<br>그저 스쳐가는 바람결이었네<br>수많은 얼굴 속 익명인 채<br>존재했지만 존재하지 않던<br><br>그러나 한 글자 한 글자 마음 실어<br>당신의 이름을 부르니<br>어둠 속 별처럼 빛나기 시작하네<br>마침내 내게 다가와<br>영혼의 꽃송이 되어 피어오르네<br><br>아껴주던 모든 것들이<br>이제 비로소 진짜 모습 드러내고<br>그리움 깊었던 시간들이<br>환영하는 눈빛 되어 돌아오네<br><br>깊은 밤 홀로 서 있을 때조차<br>누군가 알아봐 준다는 안도감<br>그것은 곧 삶의 의미이며<br>세상을 살아갈 이유였구나<br>(<The Moment I Call Your Name><br>**Before I called your name**<br>**You were merely a passing breeze**<br>**Anonymous amidst countless faces**<br>**Existing, yet not truly existing**<br><br>But as I called your name<br>With each syllable laden with my heart<br>You began to shine like a star in the darkness<br>Finally, you come to me<br>**And bloom as a flower of the soul**<br><br>All the things I cherished<br>Now finally reveal their true forms<br>And the times of deep longing<br>Return as eyes filled with welcome |
|---|---|

| | Even when standing alone in the deep night<br>The relief that someone recognizes me—<br>That was the very meaning of life<br>And the reason to live in this world) |
|---|---|
| **Top 5 Emotion Categories (KPoEM)** | **Top 5 Emotion Categories (KPoEM)** |
| care: 0.90<br>realization: 0.87<br>admiration: 0.86<br>joy: 0.79<br>expectancy: 0.78 | admiration: 0.92<br>care: 0.89<br>expectancy: 0.87<br>realization: 0.86<br>joy: 0.83 |

Table 15 presents the emotion distributions produced by the KPoEM emotion classification model for both the input poem—an excerpt from Kim Chunsu's *Kkot*—and the poem generated by the proposed RAG-based model, titled *Ireum bureuneun sungan* (Korean: 이름 부르는 순간; The Moment I Call Your Name). In both texts, prominent emotion categories such as Care, Realization, Admiration, Expectation, and Joy establish a shared affective structure. Moreover, the generated poem demonstrates high semantic coherence by effectively inheriting the input's core metaphor of naming. Consequently, this close alignment in both emotion distributions and thematic substance demonstrates the model's efficacy in preserving the input's core emotional orientation throughout the generation process.

# VI. Conclusions[14]

This study proposed a new methodology for the quantitative analysis of Korean modern poetry through the construction of KPoEM, an emotion-labeled dataset annotated at both the line and work levels. KPoEM consists of 7,662 entries, each annotated with 44 fine-grained emotion categories in a multi-label scheme by five independent annotators. The resulting dataset was then used for sequential fine-tuning of KcELECTRA, which had been initially fine-tuned on the KOTE dataset.

Quantitative evaluation on a held-out test set of 763 entries demonstrated that the proposed KPoEM model outperformed the models fine-tuned directly on the KOTE dataset and on KPoEM alone across all metrics, achieving an accuracy of 0.79, an F1 (micro) score of 0.60, and an MCC of 0.47. In qualitative evaluation, the KPoEM model accurately captured not only the dominant emotions within poems but also the contextual emotions embedded in the text, showing particularly clear recognition of emotional characteristics in poems reflecting the sentiments of the colonial period.

Nevertheless, this study has several limitations. Due to copyright constraints and historical considerations, the dataset is restricted to works by five representative poets, resulting in limited temporal and authorial diversity, as well as the underrepresentation of female poets. In addition, the inherent ambiguity of poetic language and the subjectivity of emotional interpretation remain fundamental challenges that cannot be fully resolved through computational approaches alone. Despite these limitations, this study experimentally demonstrates that the emotional structures embedded in poetic texts can be systematically explored through data-driven methods, providing foundational resources for expanding the intersection of literature and artificial intelligence.

[14] To ensure the reproducibility and extensibility of the research, all source code used in this study has been made publicly available in the following repository. See the links below for details. https://github.com/AKS-DHLAB/KPoEM

Aligned with recent advancements in LLM-based poetry generation and evaluation, KPoEM provides a practical foundation for both AI-assisted creative education and the preservation of Korean literary affect as structured data. KPoEM enables learners to intuitively grasp complex emotional layers in poetry, supporting AI-assisted creation and revision based on targeted emotional tones. By interacting with AI, learners can experientially explore the creative process while internalizing the distinctive stylistic features of Korean literature. Ultimately, KPoEM serves as a practical reference for emotion-driven interpretation and creation, facilitating the data-driven preservation of literary sensibility within the literature-AI intersection.

Furthermore, KPoEM is conceived not merely as a standalone resource, but as the foundation for a Co-Reading environment in which humans and AI collaboratively reconstruct poetic texts through color-based sensory and emotional mediation (Lim, 2025). Future work will extend this framework by constructing a dataset of sensory elements in Korean modern poetry (Ji, 2025), thereby enabling more holistic literary analysis that encompasses the two fundamental dimensions of human experience—sense and emotion.

# Appendix A. Dataset Construction Workflow

The KPoEM (Korean Poetry Emotion Mapping) dataset was constructed through a structured preprocessing and refinement workflow.[15]

Specifically, (1) criteria for poet and poem selection were first established; (2) poetic texts were collected from Wikisource; (3) the texts then underwent cleaning and normalization procedures; (4) an emotion annotation framework was applied; and finally, (5) each entry was independently labeled by five expert annotators to ensure high data reliability.

## A.1. Poet and Poem Selection Criteria

For the construction of the KPoEM, poet selection is conducted using the following three criteria. First, poets are selected based on bibliographic data of doctoral dissertations in the field of modern Korean literature published between 2000 and 2019. Specifically, poets whose names appear ten or more times in dissertation titles or subtitles are included (Kim & Cheon, 2020). This criterion ensures the representativeness of poets who have been recurrent subjects of literary scholarship. Second, additional poets are selected based on public recognition, referencing 'the Top 10 Korean Poets' list published by the Korean Poets Association (Seoul Newspaper, 2007). Third, five representative modern Korean poets—Han Yong-un (Korean: 한용운), Im Hwa (Korean: 임화), Kim So-wol (Korean: 김소월), Yi Sang (Korean: 이상), and Yun Dong-ju (Korean: 윤동주)—whose copyright restrictions have expired (i.e., deceased over 70 years ago), are chosen to ensure legal clarity for dataset publication.

The primary sources are public domain poetry collections available on Wikisource, including *Nimui chimmuk*(Korean: 님의 침묵; Silence of My Beloved) by Han Yong-un, *Hyeonhaetan* (Korean: 현해탄) by Im Hwa, *Jindallaekkot* (Korean: 진달래꽃; Azaleas) by Kim So-wol, and *Haneulgwa baramgwa byeolgwa si* (Korean: 하늘과 바람과 별과 시; Sky, Wind, Stars and Poetry) by Yun Dong-ju. In the case of Yi Sang, the dataset focuses on his Korean-language series poems such as *Widok* (Korean: 위독; *Critical Condition*), *Ogamdo* (Korean: 오감도; Crow's Eye View), and *Yeokdan* (Korean: 역단; Reverse).

A total of 483 poems are collected, comprising 165 from Kim Sowol, 112 from Yun Dong-ju, 46 from Yi Sang, 43 from Im Hwa, and 117 from Han Yong-un. Each poem is further segmented by line to support fine-grained emotional annotation, forming the basis of the constructed dataset.

[15] In the data sample examples presented in Appendix A, titles and subtitles were translated using Gemini 3 Pro, and subsequently reviewed and validated by the authors. For proper nouns, the Romanized transliteration was provided alongside the translation to preserve the original phonetics. Furthermore, as the data refinement and preprocessing procedures hold greater significance than the specific illustrative content, the main bodies of the poems were not translated.

## A.2. Data Collection from Public Sources

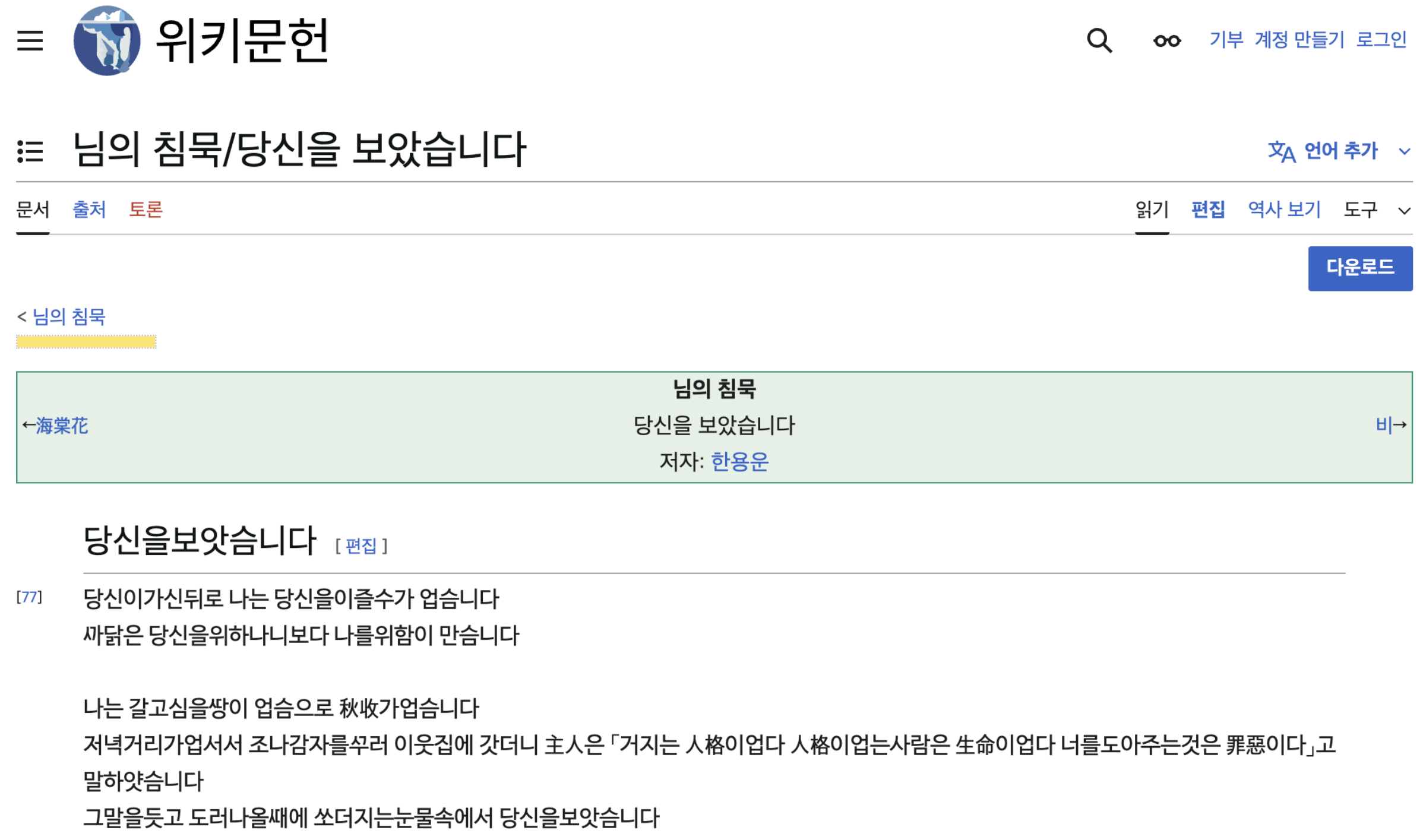

**Figure 2. Public Data Source Example: Wikisource. "*Dangsineul boatseumnida*[16]" (Korean: 당신을 보았습니다; I Saw You) by Han Yong-un.**

Wikisource, a free online digital library operated by the Wikimedia Foundation, publishes XML-formatted dump files twice a month, offering complete access to its structured content (see Figure 2). Each dump file includes metadata such as document titles, text bodies, and revision histories, making it a valuable resource for large-scale corpus construction.

In this study, we construct a poetry dataset using the Korean Wikisource (kowikisource) dump file dated November 1, 2024. The dump file[17] is converted into JSON format using the Wikiextractor tool, and each document is subsequently parsed at both the line level and the work level. The resulting JSON objects include keys such as id, revid, url, title, and text. Table 16 presents an example of raw text extracted from the Korean Wikisource before any preprocessing steps. This excerpt is from the poem *Dangsin-eul boat-seum-nida* (Korean: 당신을 보았습니다; I Saw You) by Han Yong-un.

**Table 16. Example of Raw Text Extracted from Wikisource (Before Preprocessing)**

{"id": "3647", "revid": "5442", "url": "https://ko.wikisource.org/wiki?curid=3647", "title": "님의 침묵/당신을 보았습니다", "text": "<poem>
당신이 가신 뒤로 나는 당신을 잊을 수가 없습니다

[16] Wikisource, *Dangsineul boatseumnida* (Korean: 당신을 보았습니다; I Saw You, https://ko.wikisource.org/wiki/님의_침묵/당신을_보았습니다

[17] https://dumps.wikimedia.org/kowikisource/20241101/kowikisource-20241101-pages-articles.xml.bz2 (Since Wikimedia dump files are periodically removed, this file may no longer be available. Use the latest dump file for parsing.)

까닭은 당신을 위하느니보다 나를 위함이 많습니다
나는 갈고심을 땅이 없으므로 秋收가 없습니다
저녁거리가 없어서 조나 감자를 꾸러 이웃집에 갔더니 主人은「거지는 人格이 없다 人格이 없는 사람은
生命이 없다 너를 도아주는 것은 罪惡이다」고 말하였습니다
그 말을 듣고 돌아나올 때에 쏟아지는 눈물 속에서 당신을 보았습니다
나는 집도 없고 다른 까닭을 겸하여 民籍이 없습니다
「民籍 없는 者는 人權이 없다 人權이 없는 너에게 무슨 貞操냐」하고 凌辱하려는 將軍이 있었습니다
그를 抗拒한 뒤에 남에게 대한 激憤이 스스로의 슬픔으로 化하는 刹那에 당신을 보았습니다
아아 온갖 倫理, 道德, 法律은 칼과 黃金을 祭祀 지내는 烟氣인 줄을 알았습니다
永遠의 사랑을 받을까 人間歷史의 첫 페이지에 잉크칠을 할까 술을 마실까 망서릴 때에 당신을
보았습니다</poem>"}

The converted JSON data is loaded in a Python (v3.10.18) environment and organized into a unified DataFrame structure using the pandas library. Poems by specific authors are retrieved via keyword search in the title field, and the corresponding text field is extracted as the primary content. HTML escape characters (e.g., <poem>) within the text are restored using the html.unescape function, and tags such as <poem> and </poem> are removed directly through string processing using regular expressions. The cleaned texts were subsequently split by newline characters (\n), segmenting each poem into individual lines, and were also divided into work-level versions split into 512-character segments, resulting in two structured versions of the dataset.

Each line is organized into a DataFrame with the column names line_id (unique identifier), poem_id (poem identifier), text (line content), sub_title (subtitle, if applicable), title (poem title), and poet (poet name). This initial line-level structuring of the poetic content forms the foundational layer for subsequent emotion annotation and computational analysis (see Table 17).

**Table 17. Example of Line-Level Segmentation from Processed Wikisource Data**

| line_id | poem_id | text | sub_title | title | poet |
|---|---|---|---|---|---|
| **2849** | **197** | 당신이 가신 뒤로 나는 당신을 잊을 수가 없습니다 | | 당신을 보았습니다 (I Saw You) | Han Yong-un |
| **2850** | **197** | 까닭은 당신을 위하느니보다 나를 위함이 많습니다 | | 당신을 보았습니다 (I Saw You) | Han Yong-un |
| **2851** | **197** | 나는 갈고심을 땅이 없으므로 秋收가 없습니다 | | 당신을 보았습니다 (I Saw You) | Han Yong-un |
| **2852** | **197** | 저녁거리가 없어서 조나 감자를 꾸러 이웃집에 갔더니 主人은「거지는 人格이 없다 人格이 없는 사람은 生命이 없다 너를 도아주는 것은 罪惡이다」고 말하였습니다 | | 당신을 보았습니다 (I Saw You) | Han |

| | | | | | Yong-un |
|---|---|---|---|---|---|
| **2853** | **197** | 그 말을 듣고 돌아나올 때에 쏟아지는 눈물 속에서 당신을 보았습니다 | | 당신을 보았습니다 (I Saw You) | Han Yong-un |
| **2854** | **197** | 나는 집도 없고 다른 까닭을 겸하여 民籍이 없습니다 | | 당신을 보았습니다 (I Saw You) | Han Yong-un |
| **2855** | **197** | 「民籍 없는 者는 人權이 없다 人權이 없는 너에게 무슨 貞操냐」하고 凌辱하려는 將軍이 있었습니다 | | 당신을 보았습니다 (I Saw You) | Han Yong-un |
| **2856** | **197** | 그를 抗拒한 뒤에 남에게 대한 激憤이 스스로의 슬픔으로 化하는 刹那에 당신을 보았습니다 | | 당신을 보았습니다 (I Saw You) | Han Yong-un |
| **2857** | **197** | 아아 온갖 倫理, 道德, 法律은 칼과 黃金을 祭祀 지내는 烟氣인 줄을 알았습니다 | | 당신을 보았습니다 (I Saw You) | Han Yong-un |
| **2858** | **197** | 永遠의 사랑을 받을까 人間歷史의 첫 페이지에 잉크칠을 할까 술을 마실까 망서릴 때에 당신을 보았습니다 | | 당신을 보았습니다 (I Saw You) | Han Yong-un |

### A.3. Preprocessing Pipeline

Prior to labeling, the poetry texts collected from Wikisource—both line-level and work-level—undergo additional cleaning and normalization to ensure their suitability for training emotion classification models.

1. **Text Selection Criteria and Hangeul–Sino-Korean Notation Rules**

   When poems contain both archaic Korean and modern Korean translations, only the modern version is retained in the dataset to ensure consistency and improve classification accuracy. In cases where multiple versions of a poem exist, preference is given to the version that preserves the original orthography and includes accurate notation of Sino-Korean characters (漢字). In such cases, the selected version is further normalized by researchers, who apply consistent principles to handle Hangeul–Sino-Korean pairings (e.g., 한글 [漢字]), ensuring both readability and fidelity to the poetic source (see Table 18). Additionally, if discrepancies in line order are found, the parsed data is manually realigned to match the canonical order in the source. This process ensures that each annotated line preserves its semantic position within the work.

**Table 18.** ***Tagorui sireul ikgo*** **(Korean: 타골의 시를 읽고; Reading Tagore's Poems), by Han Yong-un**

| **Preserved Original Version** | **Duplicate Version** | **Key Differences** | **Version Selected for Research Use** |
|---|---|---|---|

| 벗이여 나의 벗이여 愛人의 무덤 위의 피어 있는 꽃처럼<br>나를 울리는 벗이여<br>작은 새의 자취도 없는 沙漠의 밤에 문득 만난 님처럼<br>나를 기쁘게 하는 벗이여<br>… | 벗이여, 나의 벗이여. 애인의 무덤 위에 피어 있는 꽃처럼 나를 울리는 벗이여. 작은 새의 자취도 없는 사막의 밤에 문득 만난 님처럼 나를 기쁘게 하는 벗이여.<br>… | Preserved original orthography; accurate Sino-Korean character notation | 벗이여 나의 벗이여 애인(愛人)의 무덤 위의 피어 있는 꽃처럼<br>나를 울리는 벗이여<br>작은 새의 자취도 없는 사막(沙漠)의 밤에 문득 만난 님처럼<br>나를 기쁘게 하는 벗이여<br>… |
|---|---|---|---|

## 2. Data Exclusion

Poems written in archaic Korean script (e.g., ᄒᆞ) are excluded from the initially cleaned dataset due to incompatibilities with modern encoding environments. A representative example is Kim Sowol's *Oilbam sanbo* (Korean: 5일밤 산보; Five-Night Walk), which is removed during this filtering step. In the work-level dataset, duplicated expressions are preserved as they appear in the original text; however, in the line-level dataset, duplicate lines are identified and eliminated to prevent unnecessary redundancy. Poems with strong visual formatting or image elements—such as Yi Sang's *Ogamdo sije 4ho* and *5ho*—are also excluded, as their visual structure cannot be accurately reproduced for quantitative analysis (See Figure 3).

**烏瞰圖 詩第四號 / 오감도 시제4호** [ 편집 ]

患者의容態에關한問題.

·0 9 8 7 6 5 4 3 2 1
0·9 8 7 6 5 4 3 2 1
0 9·8 7 6 5 4 3 2 1
0 9 8·7 6 5 4 3 2 1
0 9 8 7·6 5 4 3 2 1
0 9 8 7 6·5 4 3 2 1
0 9 8 7 6 5·4 3 2 1
0 9 8 7 6 5 4·3 2 1
0 9 8 7 6 5 4 3·2 1
0 9 8 7 6 5 4 3 2·1
0 9 8 7 6 5 4 3 2 1·

診斷 0 : 1

26.10.1931

以上 責任醫師 李 箱

환자의 용태에 관한 문제.

·0 9 8 7 6 5 4 3 2 1
0·9 8 7 6 5 4 3 2 1
0 9·8 7 6 5 4 3 2 1
0 9 8·7 6 5 4 3 2 1
0 9 8 7·6 5 4 3 2 1
0 9 8 7 6·5 4 3 2 1
0 9 8 7 6 5·4 3 2 1
0 9 8 7 6 5 4·3 2 1
0 9 8 7 6 5 4 3·2 1
0 9 8 7 6 5 4 3 2·1
0 9 8 7 6 5 4 3 2 1·

진단 0 : 1

26.10.1931

이상 책임의사 이 상

**Figure 3. Public Data Source Example: Wikisource**
***"Ogamdo sije 4ho*[18]*"* (Korean: 오감도 시제 4호; Crow's Eye View No. 4) by Yi Sang.**

[18]Wikisource, *Ogamdo sije 4ho* (Korean: 오감도 시제 4호; Crow's Eye View No. 4, https://ko.wikisource.org/wiki/오감도

### 3. Character Normalization and Cleaning

Character normalization is performed to ensure consistency in textual representation. Sino-Korean words are retained alongside Hangul in the form of Hangul (Sino-Korean), e.g., 공기 (空氣), while alternative notation forms already present in Wikisource (e.g., 단[熱한]) are preserved as-is. Non-textual symbols, excessive punctuation (e.g., …… ), and formatting artifacts are retained in the work-level dataset but removed in the line-level dataset as unnecessary noise. In cases where typographical errors are identified in Wikisource, corrections are made by referencing official printed editions of the poetry collections. For example, the misspelling *sadiri* (Korean: 사디리) is corrected to the standard form *sadari* (Korean: 사다리; ladder)[19].

### 4. Preservation of Poetic Features

To maintain the literary integrity of the source material, poetic license, older expressions, and archaisms are preserved without modernizing the language. For instance, expressions such as *Ani nunmul heulli-orida* (Korean: 아니 눈물 흘리오리다; Will I not shed tears?) are retained in their original form. Repetitive notation such as 々, used to indicate duplication in Sino-Korean, is also preserved to avoid semantic distortion.

### 5. Exception Handling in Line and Sentence Segmentation

In the line-level dataset, most poetic lines are segmented based on newline characters (\n). Most poems are segmented using newline characters (\n). However, some poems are written in prose-like format without explicit line breaks. In such cases, if the text exceeds four sentences, segmentation is performed at sentence boundaries—typically identified by sentence-final endings such as *-da* (Korean: -다), *-yo* (Korean: -요), or *-orida* (Korean: -오리다)—as shown in Table 19 and Table 20, in order to preserve semantic integrity.

**Table 19. Original Poem Text**

| Original Poem Text | Sub-Title | Title | Poet |
|---|---|---|---|
| 벌판한복판에 꽃나무하나가있소. 근처(近處)<br>에는꽃나무가하나도없소.<br>꽃나무는제가생각하는꽃나무를 열심(熱心)<br>으로생각하는것처럼 열심으로꽃을피워가지고섰소.<br>꽃나무는제가생각하는꽃나무에게갈수없소.<br>나는막달아났소. 한꽃나무를위(爲)하여 그러는것처럼<br>나는참그런이상스러운흉내를내었소. | | 꽃나무 (Blossom Tree) | Yi Sang |

**Table 20. Line-Level Segmentation (Processed)**

[19] Han, Y. (2016). *Silence of My Beloved 1.* Doseochulpan Chaekkkoji.

| line_id | poe<br>m_id | text | sub-title | title | poet |
|---|---|---|---|---|---|
| 6699 | 450 | 벌판한복판에 꽃나무하나가있소. | | 꽃나무 (Blossom Tree) | Yi San<br>g |
| 6700 | 450 | 근처(近處)에는꽃나무가하나도없소. | | 꽃나무 (Blossom Tree) | Yi San<br>g |
| 6701 | 450 | 꽃나무는제가생각하는꽃나무를 열심(熱心)<br>으로생각하는것처럼<br>열심으로꽃을피워가지고섰소. | | 꽃나무 (Blossom Tree) | Yi San<br>g |
| 6702 | 450 | 꽃나무는제가생각하는꽃나무에게갈수없소. | | 꽃나무 (Blossom Tree) | Yi San<br>g |
| 6703 | 450 | 나는막달아났소. | | 꽃나무 (Blossom Tree) | Yi San<br>g |
| 6704 | 450 | 한꽃나무를위(爲)하여 그러는것처럼<br>나는참그런이상스러운흉내를내었소. | | 꽃나무 (Blossom Tree) | Yi San<br>g |

## A.4. Emotion Annotation Framework

### 1. Emotion Label Taxonomy

To capture the nuanced and compound emotional expressions in modern Korean poetry, we adopt a multi-label annotation scheme based on the KOTE dataset (Jeon et al., 2024). This taxonomy comprises 44 fine-grained emotion categories, refined and clustered from Korean emotion vocabulary using word embeddings and human validation (see Table 1). The emotion labels span a wide range of affective states—including primary emotions (e.g., anger, joy, sadness), social or relational states (e.g., care, welcome, respect), and culturally specific expressions (e.g., *bijang-ham* [resolute], *seoreo-um* [sorrow]). Each poetic line can be annotated with up to 10 concurrent emotion tags, acknowledging the inherent emotional multiplicity in poetic expression.

### 2. Controlling Contextual Dependency via Shuffling

In the line-level dataset, to empirically test whether individual poetic lines contain sufficient lexical cues for emotion classification independent of broader poetic context, we constructed a shuffled dataset and carried out annotation on it (see Table 21). In this dataset, the original order of lines within each poem was randomly permuted, effectively removing the narrative context and structural continuity that could otherwise influence emotional interpretation.

**Table 21. A Randomly Shuffled Representative Sample from the Final Line-Level Annotated KPoEM**

| line_id | poem_id | text | sub_title | title | poet | annotator_01 | annotator_02 | annotator_03 | annotator_04 | annotator_05 |
|---|---|---|---|---|---|---|---|---|---|---|
| 6999 | 483 | 장난감신부에게 내가 바늘을주면 장난감신부는 아무것이나 막 찌른다.일력. 시집. 시계. 또 내몸 내 경험이들어앉아있음직한곳. | 2밤 (Night 2) | I WED A TOY BRIDE | Yi Sang | anxiety, fear, guilt | fear, surprise, embarrassment, sadness, realization | fear, embarrassment, reluctant, dissatisfaction, disappointment, preposterous, contempt | shock, fear, anxiety, surprise, embarrassment, contempt | shock, fear, surprise, embarrassment, anxiety, despair, gessepany, exhaustion |
| 6930 | 480 | 천고로창천이허방빠져있는함정에유언이석비처럼은근히침몰되어있다. | 자상 (Self-Injury) | 위독 (Critical Condition) | Yi Sang | despair, shock, anxiety, surprise, fear | despair, gessepany, exhaustion, realization, sadness | embarrassment, reluctant, disappointment | shock, fear, despair, gessepany, exhaustion | disappointment, despair, compassion, anxiety, sadness |
| 4760 | 364 | 삼수갑산이 어디뇨 내가 오고 내 못 가네 | | 삼수갑산 - 차안서삼갑산운 (Samsugapsan - Chaanseosamgapsanun) | Kim So-wol | sadness, expectancy, disappointment, sorrow, anxiety, dissatisfaction, exhaustion | embarrassment, dissatisfaction, sorrow, sadness, disappointment, preposterous, despair, fed up, pathetic | embarrassment, disappointment, sorrow, dissatisfaction | embarrassment, anxiety, sorrow, sadness | embarrassment, pathetic, exhaustion, gessepany, guilt |

This experimental design draws on the structural premise of the KOTE dataset. KOTE is built to classify emotions embedded in single, self-contained comments under 512 characters, without relying on surrounding textual context. Similarly, we treat each poetic line as an atomic unit, equivalent to a comment, and create an annotation environment in which raters assess the emotional content of a line based solely on its intrinsic lexical and stylistic features, isolated from neighboring lines.

This randomized arrangement serves two primary purposes:

- It enables a focused analysis of intra-line emotional signals by eliminating inter-line dependencies; and
- It encourages emotion classification models to rely on semantic properties of the line itself, rather than on positional or narrative flow cues.

### 3. Annotation Process

To ensure reliability and consistency, five annotators participated in labeling emotions for the same dataset. All annotators were trained researchers specializing in Korean literature and digital humanities, and each independently labeled the dataset in separate annotation sheet environments without interference from one another. In cases where conflicts arose in the annotation results, a third-party reviewer cross-validated the data and adjudicated disagreements.

To minimize subjectivity, all annotators received prior training and were provided with a standardized guideline document. This guideline included: (1) operational definitions of each emotion label, (2) illustrative labeling examples prepared by the author, (3) explicit instructions to label the work-level dataset only after completing the line-level dataset in order to remain faithful to the dataset's purpose, and (4) clear guidance on assigning multiple emotion labels. Annotators were encouraged to assign multiple emotions—up to a maximum of ten per line—whenever semantically justified. This approach follows the latest standard procedures for multi-annotator human annotation studies in NLP (Jeon et al., 2024).

# Appendix B. Prompt Design for Emotion-Aware Poetry Generation

This appendix provides the specific prompt template used for the generative phase of the framework. The generation process utilizes the KPoEM dataset and a specialized Emotion Classification Model to ensure that the output maintains both poetic quality and emotional consistency.

## B.1. Prompt Structure and Input Variables

The prompt is structured using a PromptTemplate that integrates three core components (See Figure 1):

- **sample_text (Input Poem):** The original source poem provided by the user.
- **context_snippets (Augmented Context):** 10 poetic segments retrieved from the KPoEM vector database. These snippets are selected through emotion-constrained filtering to provide stylistic and lexical inspiration aligned with the target emotion.
- **top_emotion (Emotion Scores):** An array of emotion classification scores generated by the KPoEM model. These scores act as the foundational benchmark to guide the tone of the generated poem.

## B.2. Generative Prompt Template

The following template is used to instruct the Large Language Model (LLM) to act as a creative modern poet (See Table 22):

**Table 22. Generative Prompt Template: Original Korean Text and English Translation**

| Original Prompt (Korean) |
|---|
| PromptTemplate(<br>input_variables=["sample_text", "context_snippets", "top_emotion"],<br>template="""<br>### 시스템:<br>당신은 창의적이고 감성적인 근현대 시인입니다.<br>아래 조건을 모두 만족하는 시를 작성하세요.<br><br>- 한국어로만 작성할 것<br>- 이모지나 그림 없이 순수 시어만 사용할 것<br>- 원문 텍스트는 시의 소재·상징·분위기 아이디어로 활용<br>- 참고 문장들에서는 표현 방식과 어휘의 결을 배우되 그대로 베끼지 말 것<br>- 감정 목록은 시 전체의 정서적 축으로 삼을 것 |

```
- 제목을 지을 것
- 해설 금지, 시만 작성
---
**원문 텍스트:**
{sample_text}
---
**참고 문장들:**
{context_snippets}
---
**핵심 감정 목록:**
{top_emotion}
---

### 시:
""".strip())
```

**English Translation (Gemini 3 Pro)**

```
PromptTemplate(
    input_variables=["sample_text", "context_snippets", "top_emotion"],
    template="""
    ### System:
    You are a creative and emotional modern Korean poet.
    Write a poem that satisfies all of the following conditions:

    - Must be written in Korean only.
    - Use only pure poetic language (no emojis or images).
    - Use the Source Text as a conceptual idea for the subject, symbolism, and atmosphere.
    - Learn the expression style and nuance from the Reference Sentences, but do not copy them directly.
    - Use the Core Emotion List as the affective axis for the entire poem.
    - Create a title for the poem.
    - Do not provide an explanation; write the poem only.
    ---
    **Source Text:**
    {sample_text}
    ---
    **Reference Sentences:**
    {context_snippets}
    ---
    **Core Emotion List:**
```

```
{top_emotion}
---

### Poem:
""".strip())
```